\pdfoutput=1
\documentclass[11pt]{article}

\usepackage[final]{acl}
\usepackage{verbatim}
\usepackage{listings}
\usepackage{longtable}            
\usepackage{afterpage}
\usepackage{array} 
\usepackage{booktabs} 
\usepackage{times}
\usepackage{latexsym}
\usepackage{amsmath}
\usepackage{pdfpages}
\usepackage[T1]{fontenc}
\usepackage[utf8]{inputenc}

\usepackage{microtype}

\usepackage{inconsolata}

\usepackage{graphicx}

\title{BPP-Search: Enhancing Tree of Thought Reasoning for Mathematical Modeling Problem Solving}

\author{
 \textbf{Teng Wang\textsuperscript{1}},
 \textbf{Wing-Yin Yu\textsuperscript{2}},
 \textbf{Zhenqi He\textsuperscript{1}},
 \\
 \textbf{Zehua Liu\textsuperscript{2}},
 \textbf{Hailei Gong\textsuperscript{2}},
  \textbf{Han Wu\textsuperscript{2}},
  \textbf{Xiongwei Han\textsuperscript{2}},
 \\
 \textbf{Wei Shi\textsuperscript{2}},
 \textbf{Ruifeng She\textsuperscript{2}},
 \textbf{Fangzhou Zhu\textsuperscript{2}},
 \textbf{Tao Zhong\textsuperscript{2}}
\\
 \textsuperscript{1}the University of Hong Kong, \\
 \textsuperscript{2}Noah's Ark Lab, Huawei
\\
wt0318@connect.hku.hk
}
\begin{document}
\maketitle

\begin{abstract}

LLMs exhibit advanced reasoning capabilities, offering the potential to transform natural language questions into mathematical models. However, existing open-source datasets in operations research domain lack detailed annotations of the modeling process, such as variable definitions, focusing solely on objective values, which hinders reinforcement learning applications. 
To address this, we release the \textbf{StructuredOR} dataset, annotated with comprehensive labels that capture the complete mathematical modeling process. We further propose \textbf{BPP-Search}, an algorithm that integrates reinforcement learning into a tree-of-thought structure using \textbf{B}eam search, a \textbf{P}rocess reward model, and a pairwise \textbf{P}reference algorithm. This approach enables efficient exploration of tree structures, avoiding exhaustive search while improving accuracy. 
Extensive experiments on StructuredOR, NL4OPT, and MAMO-ComplexLP datasets show that BPP-Search significantly outperforms state-of-the-art methods.
In tree-based reasoning, BPP-Search excels in accuracy and efficiency, enabling faster retrieval of correct solutions.
%
% The StructuredOR dataset is available at \href{https://github.com/LLM4OR/StructuredOR}{https://github.com/LLM4OR/StructuredOR}.
The StructuredOR dataset is available on Huggingface\footnote{\url{https://huggingface.co/datasets/LLM4OR/StructuredOR}} and GitHub\footnote{\url{https://github.com/LLM4OR/StructuredOR}}.

\end{abstract}

% Process reward model
% \\
% Outcome reward model
% \\
% ToT, CoT, SC
% \\
% MCTS
\section{INTRODUCTION}

Mathematical modeling, particularly Linear Programming (LP) and Mixed Integer Programming (MIP), plays a critical role in industrial applications such as logistics~\cite{logistics}, electricity scheduling and transmission~\cite{electricity}, and supply chain management~\cite{supply_chain}. With the advent of Large Language Models (LLMs), transforming natural language questions into mathematical models has become a promising approach for automating operations research tasks~\cite{MLPrompt,complexOR,orlm, hrm}. 

Despite the increasing availability of open-source operations research (OR) datasets designed for question-to-model transformation~\cite{complexLP,nl4opt,industryOR}, these datasets primarily focus on objective values while lacking detailed annotations of the underlying modeling processes. This gap limits the application of Reinforcement Learning (RL), as prior studies~\cite{prm_openai,prm_google,openai-verifier-small-model} have shown that process information can significantly enhance mathematical reasoning performance. To address this limitation, we design a rigorous framework for dataset generation and introduce the \textbf{StructuredOR} dataset, which not only provides objective values for evaluation but also includes comprehensive annotations of the modeling process, enabling broader applicability in RL-based methods.

Chain-of-Thought (CoT)~\cite{cot}, Self-Consistency (SC)~\cite{sc} and Tree-of-Thought (ToT)~\cite{tot} have demonstrated substantial improvements in reasoning tasks. However, these approaches have inherent limitations. CoT heavily relies on the policy model and generates only one reasoning path at a time, making it likely to fail to find the correct answer when the policy model is weak. SC, without a verifier, struggles to validate the correctness of candidate answers, allowing errors in intermediate steps to propagate and mislead the reasoning process. Similarly, ToT generates multiple leaf nodes as potential answers, but without a verifier, it is unclear which leaf node should be selected as the final solution. Nonetheless, ToT remains promising; with sufficiently wide and deep trees and effective node selection strategies, it has the potential to generate optimal solutions.

To enhance the reasoning process within the ToT framework, we propose \textbf{BPP-Search}, a novel method that integrates \textbf{B}eam Search, a \textbf{P}rocess Reward Model (PRM), and a pairwise \textbf{Preference} algorithm. BPP-Search is designed to improve accuracy and reduce unnecessary node exploration, making it particularly effective for complex reasoning tasks in mathematical modeling.

Our contributions are threefold: (1) We introduce the \textbf{StructuredOR} dataset, which bridges the gap between existing datasets and the requirements of RL-based methods by providing detailed modeling annotations. (2) We propose \textbf{BPP-Search} and explore heuristic algorithms combined with PRM, including Beam Search~\cite{beam-search}, Greedy~\cite{greedy}, Epsilon Greedy~\cite{epsilon_greedy}, and Random Greedy we proposed in Section~\ref{sec:method:random_greedy}. (3) We conduct extensive experiments on the \textbf{StructuredOR}, \textbf{NL4OPT}~\cite{nl4opt}, and \textbf{Mamo-ComplexLP}~\cite{complexLP} datasets, demonstrating the superiority of BPP-Search over baseline and current state-of-the-art methods from the perspective of efficiency and accuracy.

\section{RELATED WORK}

\subsection{Mathematical Modeling Datasets}
Mathematical modeling datasets can be broadly categorized into two types: abstract modeling and concrete instance modeling. Modeling tools such as Pyomo~\cite{pyomo} and AMPL~\cite{ampl}, and OPL~\cite{opl} provide support for both approaches, enabling users to work with abstract models as well as concrete instances.

% Abstract modeling~\cite{complexOR} focuses on high-level representations without specific parameter values. As the size of parameters grows significantly, it becomes challenging for LLMs to generate and manage such large quantities in an orderly manner. Therefore, abstract modeling aims to extract the essential structural information of the model. For instance, MLPrompt~\cite{MLPrompt} uses abstract models to generate parameter distributions, which are then populated with specific parameter values to create concrete instances. Additionally, several works~\cite{tsp,car-pooling} leverage abstract models with CoT and LLMs to address problems such as the Traveling Salesman Problem ~\cite{tsp} and carpooling~\cite{car-pooling} without relying on traditional mathematical solvers.

Abstract modeling focuses on capturing the essential structural information of a mathematical model. It typically involves two steps: defining basic model declarations and applying data to create concrete instances. This approach is particularly suited for large-scale industrial applications and research, as models can be defined once and reused by importing different datasets. For instance, MLPrompt~\cite{MLPrompt} leverages abstract models to generate parameter distributions, which are subsequently populated with specific values to construct concrete instances within industrial pipelines. Additionally, several studies~\cite{googe-tsp,car-pooling} combine abstract models with CoT and LLMs to address problems such as Traveling Salesman Problem~\cite{tsp}, bypassing the need for traditional mathematical solvers or explicit concrete models.

% Concrete instance modeling, on the other hand, focuses on detailed representations with explicit parameter values~\cite{complexLP,industryOR,nl4opt}. This approach is particularly advantageous when certain constraints cannot be generalized into an abstract format, as specific instances allow for more precise and tailored solutions. It can be seen as a combination of abstract modeling and Named Entity Recognition ~\cite{nl4opt_result}, where numerical values are directly extracted and mapped to parameters. For smaller-scale models, concrete instance modeling avoids the two-step process of abstract modeling, reducing error accumulation across tasks~\cite{two-step}.
Concrete modeling requires all data to be available before model processing begins, making it a straightforward and efficient approach. It is particularly suited for analytical projects. This approach is especially advantageous when certain constraints are difficult or time-consuming to generalize into an abstract format, as it allows for more precise and tailored solutions, significantly reducing processing time. 
For smaller-scale mathematical models, tasks can be solved directly without treating them as a combination of abstract modeling and Named Entity Recognition~\cite{ner}, which involves first building an abstract model and then mapping numerical parameter values to it. This approach minimizes error accumulation across tasks~\cite{two-step}.

% \subsection{Process Reward Model}
% Several works~\cite{openai-verifier-small-model,prm_openai,prm_google} first introduced the concept of the Process Reward Model (PRM) and demonstrated that small-scale PRMs can significantly enhance the performance of weak and small-scale policy models. 
% Initially, PRM training relied on manually labeled data~\cite{prm_openai}. However, with the increasing demand for process labels, methods based on Monte Carlo Tree Search (MCTS)~\cite{mcts-1,mcts-2,mcts-3,mcts-4,mcts-5} were developed to simulate and assign scores to reasoning processes. 
\subsection{Process Reward Model}
Several works~\cite{openai-verifier-small-model,prm_openai,prm_google} introduce the concept of the Process Reward Model (PRM), demonstrating its ability to significantly enhance the performance of weak and small-scale policy models. Compared to the Outcome Reward Model (ORM), PRM achieves better performance but incurs higher labeling costs~\cite{prm_google}.
 Initially, PRM training relied on manually labeled data~\cite{prm_openai}. To address the growing demand for processing labels, Monte Carlo Tree Search (MCTS)-based methods~\cite{mcts-1,mcts-2,mcts-3,mcts-4,mcts-5, hrm} were later developed to simulate and assign scores to reasoning processes.
%
% While effective, MCTS-based approaches require wide and deep trees to generate labeled data through extensive rollouts, resulting in high computational costs and probabilistic outcomes. These methods demand numerous iterations for score convergence at intermediate nodes, posing significant resource challenges. 
While effective, MCTS-based approaches require wide and deep trees to generate labeled data through extensive rollouts for score convergence at intermediate nodes, resulting in extremely high computational resource demands.
In contrast, manually labeled data is deterministic and directly reflects the intended reasoning process without relying on approximations.

\section{Dataset Generation}
\label{sec:method}
\subsection{Preliminary}
\label{sec:method:preliminary}
\paragraph{Greedy.} The Greedy algorithm selects the candidate with the highest score at each step, focusing entirely on exploitation without exploration. The selection process can be formalized as:
\begin{equation}
a^* = \arg\max_{a \in A} P(a),
\end{equation}
where \(P(a)\) represents the score of candidate \(a\), and \(A\) denotes the set of candidates. 

\paragraph{Epsilon Greedy.} The Epsilon Greedy algorithm balances exploration and exploitation during candidate selection. At each step, with a probability of \(\epsilon\), the algorithm selects a candidate randomly. Otherwise, it selects the candidate with the highest score. The selection process can be formalized as:
\begin{equation}
a^* = 
\begin{cases} 
\text{random choice,} & w.p. ~~~\epsilon, \\
\arg\max_{a \in A} P(a), & w.p. ~~~ 1 - \epsilon,
\end{cases}
\end{equation}
where \(P(a)\) represents the score of candidate \(a\), and \(A\) denotes the set of candidates. The parameter \(\epsilon\) controls the probability of exploring randomly versus exploiting the best-known option.

\paragraph{Beam Search.} Beam Search is a heuristic search algorithm that explores a fixed number (\(k\)) of the most promising candidates (beam width) at each step. Unlike Greedy, which selects only the best candidate, Beam Search maintains a set of top \(k\) candidates to balance exploration and exploitation. The selection process can be formalized as:
\begin{equation}
B_{t+1} = \text{Top-}k\left(\bigcup_{a \in B_t} \text{Expand}(a)\right),
\end{equation}
where \(B_t\) represents the set of beam candidates at step \(t\), \(\text{Expand}(a)\) denotes the set of all possible successors of candidate \(a\), and \(\text{Top-}k\) selects the \(k\) candidates with the highest scores. The parameter \(k\) controls the trade-off between computational cost and search completeness.

\paragraph{Tree of Thought.}

\begin{figure}[t!]
    \centering
    \includegraphics[width=1\linewidth]{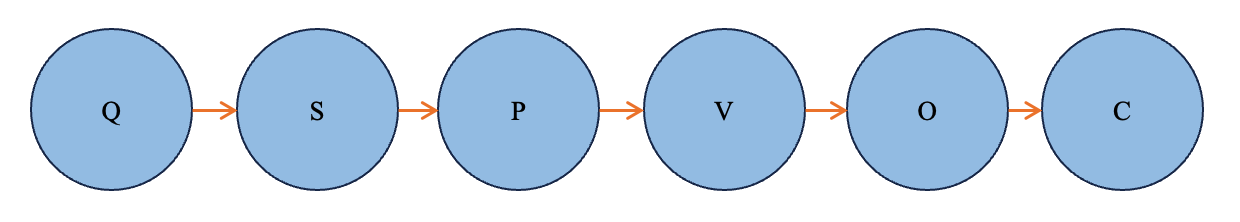}
    \caption{
Reasoning steps. The process follows the path \(Q \to S \to P \to V \to O \to C\), where \(Q\), \(S\), \(P\), \(V\), \(O\) and \(C\) represent the question, set, parameter, variable, objective, and constraint respectively.
}

    \label{fig:tot_path}
\end{figure}

\begin{figure}[t!]
    \centering
    \includegraphics[width=1\linewidth]{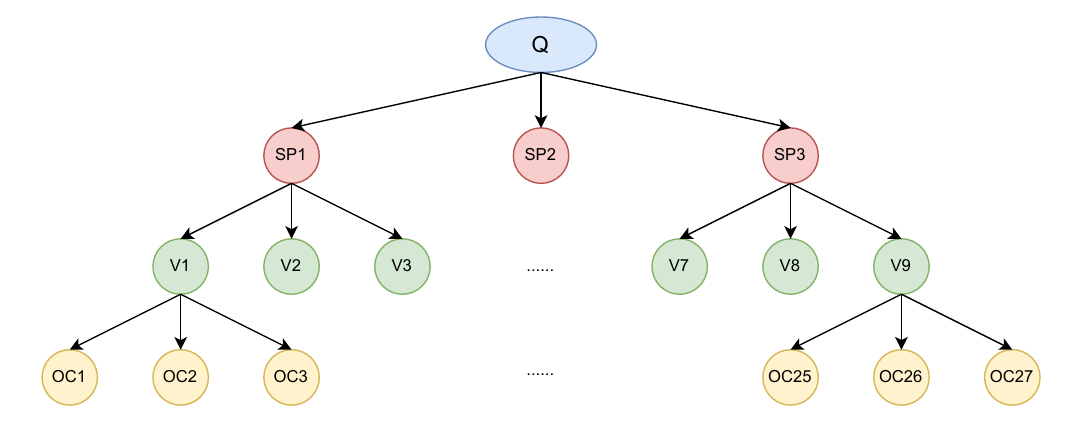}
    \caption{The structure of the Tree of Thought. Here, Q represents the question, SP represents set and parameter, V represents variable, and OC represents objective and constraint.}
    \label{fig:tot}
\end{figure}

Compared with CoT and SC, ToT has the potential to generate many accurate results, but it faces challenges in selecting a single answer from numerous leaf nodes. Our objective is to accelerate this process and efficiently retrieve a satisfactory solution. 
In the mathematical modeling task, Fig.~\ref{fig:tot_path} illustrates the reasoning process, following a structured path through the question, sets, parameters, variables, objectives, and constraints (see Appendix~\ref{appendix:reasoning_example} for an example). Constructing a six-layer tree for each example across all datasets incurs high computational costs. To address this, we group nodes based on property similarities and limit each node to a maximum of three child nodes. The resulting tree structure is as follows: the first layer represents the question, the second combines sets and parameters, the third includes variables, and the fourth integrates objectives and constraints, as shown in Fig.~\ref{fig:tot}. This approach balances tree width and computational efficiency for experiments.

\label{sec:tot}

\subsection{StructuredOR Dataset Framework}
\label{sec:structuredOR}

\begin{figure*}[t!]
    \centering
    \includegraphics[width=0.85\linewidth]{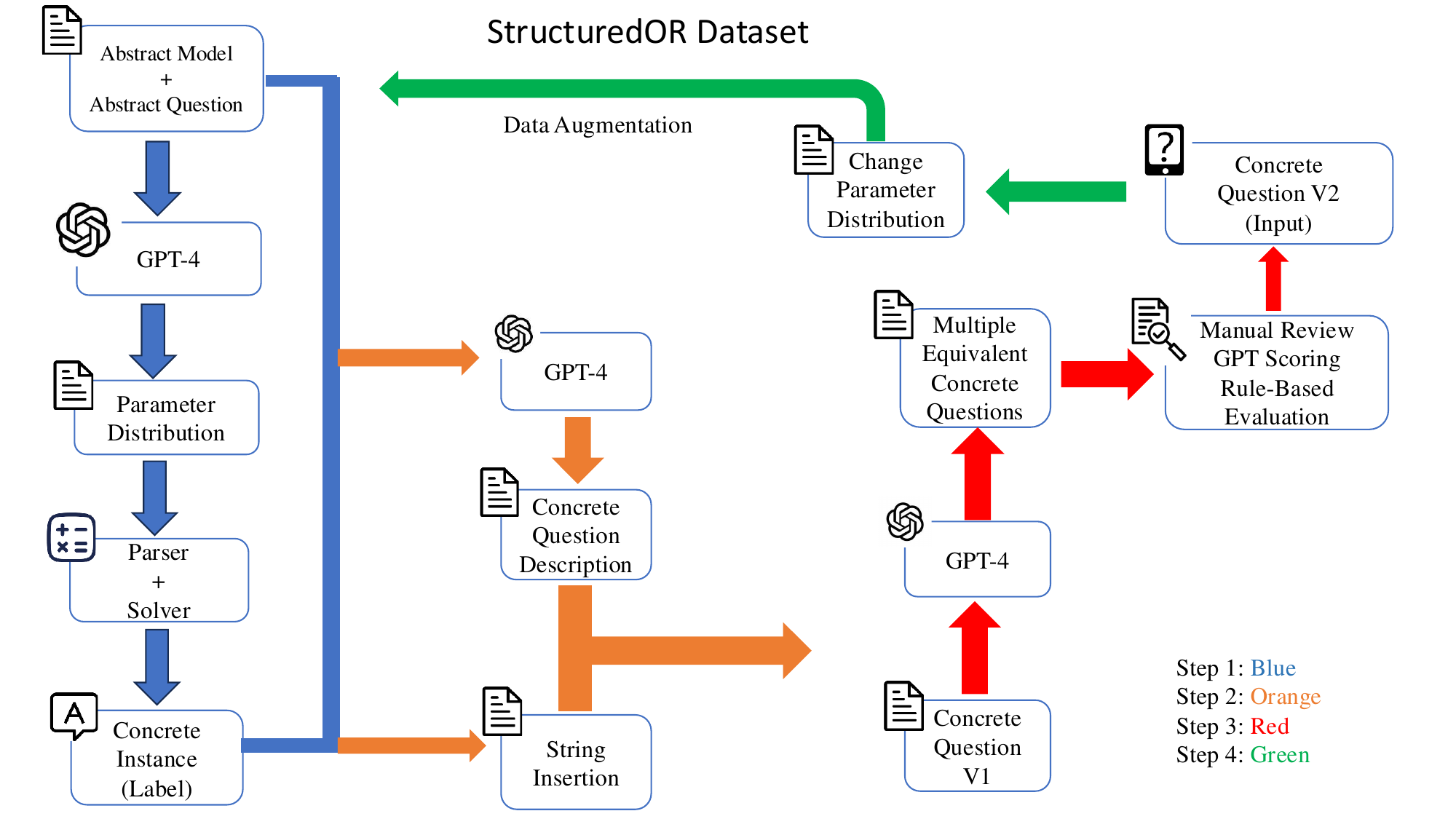}

    \caption{Pipeline of the construction process of our proposed StructuredOR dataset.}
    \label{fig:structuredOR_framework}
\end{figure*}

Existing operations research datasets predominantly focus on objective values and the annotations of the underlying modeling process appear to be missing. To bridge the gap, 
% To address the lack of annotations in existing operations research datasets
we introduce \textbf{StructuredOR}
% \footnote{The dataset will be open-sourced after the blind review process.},
a new dataset explicitly designed to provide the objective value for evaluation and capture the complete mathematical modeling process.
% This dataset expands the scope of existing resources and provides comprehensive labels, making it applicable across diverse domains.

\begin{table}[t]
\centering
\label{tab:abstract_stats}
\footnotesize
\resizebox{\linewidth}{!}{
\begin{tabular}{lcccc}
\toprule
            & Set   & Parameter & Variable & Constraint \\
\midrule
Average              & 2.22  & 3.95      & 1.40     & 2.38       \\
Max                  & 4     & 7         & 4        & 4          \\
Standard Deviation   & 0.66  & 1.07      & 0.70     & 0.85       \\
\bottomrule
\end{tabular}
}
\caption{Descriptive statistics for the abstract modeling components in the StructuredOR dataset. The table reports the number of sets, parameters, variables, and constraints in the abstract model (note that these numbers will expand after the model is instantiated into a concrete form).}

\label{tab:stat_abstract_model}
\end{table}

% Building on Xiao's work~\cite{complexOR}, which introduced the ComplexOR~\cite{complexOR} dataset for abstract modeling problems, we refine and extend this approach to encompass a broader range of abstract modeling tasks. 
% Building on the existing ComplexOR~\cite{complexOR} dataset for abstract modeling problems, we refine and extend ComplexOR to encompass a broader range of abstract modeling tasks. 
% Building on Xiao's work~\cite{complexOR}, 
Building on ~\citet{complexOR}'s work, 
which introduces a framework for generating abstract mathematical models, we refine and expand this approach to cover a wider spectrum of abstract models. 
We leverage LLM such as GPT-4o~\cite{gpt4} in conjunction with mathematical solvers to instantiate abstract models into concrete examples. Furthermore, we implement a series of validation mechanisms to construct and verify the accuracy of these concrete problems, thereby ensuring the dataset's quality and reliability.
In summary, the StructuredOR dataset provides pairs of concrete questions and corresponding model data, accompanied by comprehensive annotations detailing the entire modeling process.
Fig.~\ref{fig:structuredOR_framework} illustrates the construction pipeline, with each step distinguished by a color-coded arrow: blue for Step 1, orange for Step 2, red for Step 3, and green for Step 4. The whole process is delineated as follows:

Firstly, we leverage LLMs such as GPT-4o to generate distributions for sets and parameters in abstract models, inspired by prior work~\cite{MLPrompt} demonstrating that LLMs, such as GPT-4o, are capable of producing realistic distributions for such applications. This is followed by a simulation process to generate instance-specific parameter data.
Subsequently, these parameters are converted into concrete models in LP format using a parser
% \footnote{The parser will be open-sourced after the blind review process.}
and a modeling tool. 
The resulting models are then validated for solvability and correctness using the Gurobi solver~\cite{gurobi}.
Solvable problems from the solver are then selected as labeled instances, encompassing both the modeling process and the associated objective values. Details about standardizing the mathematical modeling data format are provided in Appendix~\ref{appendix:convention}.

% Using a parser\footnote{The parser will be open-sourced after the blind review process.} and a modeling tool, these parameters are transformed into concrete models in LP format and verified for solvability and validity using the Gurobi solver~\cite{gurobi}.
%
% Successfully solved problems are selected as labels, capturing both the modeling process and the corresponding objective values. Details about standardizing the mathematical modeling data format are provided in Appendix~\ref{appendix:convention}.

Next, based on the descriptions of sets and parameters in abstract models, we construct templates to generate markdown-formatted lists via string insertion to describe the information of sets and parameters. We subsequentially employ GPT-4o to develop a concrete problem description in natural language, which does not involve specific values for sets or parameters but provides a contextualized description of the problem. By concatenating this description with the numerical details of sets and parameters, a complete problem statement is produced. This string-based insertion ensures consistency and accuracy in the generated problem descriptions.

\begin{table}[t]
\centering

\footnotesize
\resizebox{\linewidth}{!}{
\begin{tabular}{lc}
\toprule
\textbf{Concrete Variables per Abstract Variable} & \textbf{Frequency} \\
\midrule
1 & 70  \\
2 & 214 \\
3 & 36  \\
4 & 143 \\
6 & 5   \\
8 & 22  \\
\bottomrule
\end{tabular}
}
\caption{Distribution of the number of instantiated variables derived from each abstract variable in the concrete model.}

\label{tab:values_per_variable}
\end{table}

We further utilize GPT-4o to rephrase each generated question into three semantically equivalent versions to enhance fluency and naturalness. This process is supported by a rigorous review framework, comprising manual filtering, GPT-based scoring, and rule-based evaluation, to ensure semantic consistency between the rephrased texts and their corresponding labels.
% To further improve fluency and naturalness, we employ GPT-4o to rephrase each generated question into three equivalent versions.
%
% A rigorous review process, including manual filtering, GPT-based scoring, and rule-based evaluation, ensures that the generated questions align with their corresponding labels.

% \begin{table}[t]
% \centering
% \footnotesize
% \resizebox{\linewidth}{!}{
% \begin{tabular}{lc}
% \toprule
% Instance Category             & Count \\
% \midrule
% Agriculture          & 24 \\
% Logistics            & 20 \\
% Education            & 15 \\
% Sports               & 12 \\
% Military             & 12 \\
% Energy               & 12 \\
% Telecommunications   & 11 \\
% Manufacturing        & 9  \\
% Health Services      & 5  \\
% Finance and Banking  & 4  \\
% \bottomrule
% \end{tabular}
% }
% \caption{Category Distribution of the StructuredOR Dataset}
% \label{tab:category_distribution}

% \end{table}
\begin{table*}[t]
\centering
\footnotesize
\resizebox{\linewidth}{!}{
\begin{tabular}{c|c|c|c|c|c|c|c|c|c|c}
\toprule
Category & Agriculture & Logistics & Education & Sports & Military & Energy & Telecommunications & Manufacturing & Health Services & Finance \\
\hline
% \midrule
Count & 24          & 20        & 15        & 12     & 12       & 12     & 11               & 9             & 5             & 4 \\
\bottomrule
\end{tabular}
}
\caption{Instance category distribution in the StructuredOR dataset, demonstrating broad applicability across industries.}
\label{tab:category_distribution}
\end{table*}

Finally, we introduce an iterative strategy for data augmentation by changing parameter data distributions during the initial generation phase.
% Finally, we apply data augmentation by adjusting parameter data distributions in the initial generation step and iterating the process. 
%
This yields a dataset comprising 124 concrete questions and their corresponding models, spanning domains including logistics, scheduling, and networks, of which 77 examples are from the original framework, and 47 are generated through data augmentation.
Table~\ref{tab:stat_abstract_model} presents the statistical distribution of the key components—sets, parameters, variables, and constraints—in the abstract model of the StructuredOR dataset. Table~\ref{tab:values_per_variable} summarizes the number of concrete variables instantiated from each abstract variable, while table~\ref{tab:category_distribution} shows the industry distribution.
Together, these tables illustrate the dataset's structural diversity and comprehensive domain coverage.

% Appendix~\ref{appendix:OR_dataset} provides an example of a concrete question along with its structured modeling process as the label. It also discusses the limitations of other datasets, such as Mamo-ComplexLP~\cite{complexLP}, NL4OPT~\cite{nl4opt}, and IndustryOR~\cite{industryOR}, highlighting their challenges in addressing the complete modeling process.

% 由于除了StructuredOR之外的数据集可能不适配the schema we defined in appnedix(~\ref{appendix:convention}), many of question in other datasets cannot model.So that which will leading that the rate that can successfully model is low in other dataset. Due to industryOR dataset has comparable low quantity of question, we only use structredOR, Mamo-complexLP, and NL4OPT dataset to further experiment.
Appendix~\ref{appendix:OR_dataset} provides an example of a concrete question along with its structured modeling process as the label. It also discusses the limitations of other datasets, such as Mamo-ComplexLP~\cite{complexLP}, NL4OPT~\cite{nl4opt}, and IndustryOR~\cite{industryOR}, highlighting their challenges in addressing the complete modeling process. 
%
% Since most datasets, apart from StructuredOR, are incompatible with the schema defined in Appendix~\ref{appendix:convention}, many questions in these datasets cannot be successfully parsed, resulting in lower success rates. Given the small size of the IndustryOR dataset, our experiments focus on the StructuredOR, Mamo-ComplexLP, and NL4OPT datasets.
Since most datasets, apart from StructuredOR, are incompatible with the schema defined in Appendix~\ref{appendix:convention}, some questions in these datasets cannot be successfully parsed, rendering some examples unusable. Given the small size of the IndustryOR~\cite{industryOR} dataset, our experiments focus on the StructuredOR, Mamo-ComplexLP, and NL4OPT datasets.

\subsection{PRM Dataset Preparation}
\label{sec:method:prm_dataset}

To implement process supervision within the tree structure, we introduce the PRM~\cite{prm_google, prm_openai}, which assigns a score to each intermediate step in the reasoning process.
There are two main approaches to generating training data for PRMs.
\citet{prm_google, prm_openai} rely on manually labeling each intermediate step while \citet{mcts-1,  mcts-2, mcts-3, mcts-4,  mcts-5} leverage MCTS to assign scores to intermediate steps.

% The first approach relies on manually labeling each intermediate step~\cite{prm_google, prm_openai}. The second approach leverages MCTS to assign scores to intermediate steps~\cite{mcts-1,  mcts-2, mcts-3, mcts-4,  mcts-5}.
% , particularly in scenarios where no labeled data is available.

% MCTS-based methods rely on wide and deep trees to generate labeled process data through extensive rollouts, which makes them computationally expensive. Such methods require numerous iterations to ensure score convergence for intermediate nodes, posing significant resource demands. In contrast, manually labeled data is deterministic and directly captures the intended reasoning process without relying on approximations.
MCTS-based methods rely on wide and deep trees to generate labeled process data through extensive rollouts, ensuring score convergence for intermediate nodes but demanding significant computational resources and incurring Reward Hacking~\cite{lilianweng}. In contrast, manually labeled data is deterministic and directly captures the intended reasoning process without relying on approximations.

% Due to the limitations of the policy model, parser and modeling tool capabilities, 
We first utilize the CoT and ToT frameworks across various policy models, including GPT~\cite{gpt4}, LLama~\cite{llama31}, and Qwen series~\cite{qwen25-math}, applied to the NL4Opt~\cite{nl4opt} and MAMO-ComplexLP~\cite{complexLP} training datasets. 
% This process identifies solvable examples by comparing their objective values. 
Examples with consistent objective values are assumed to have correct modeling processes and therefore do not require manual labeling.
Although the StructuredOR dataset already includes detailed modeling process annotations and serves as a source of high-quality positive examples, we adopt the same procedures to further augment its training data.

The process of modeling labels for both correct and incorrect reasoning paths is structured in alignment with the layers of the ToT framework. This follows a cumulative approach, where each layer is constructed sequentially, building upon the outcomes of the preceding layer.
% The process modeling labels for both correct and incorrect reasoning paths are segmented according to the layers of the ToT structure, following a cumulative process where each layer builds upon the previous one.
%
% If the original label represents a correct reasoning path, all segmented labels derived from it are also correct, as the correctness of each segment ensures the validity of the entire path.
If the generated label represents a correct reasoning path, all segmented labels derived from it are also correct, as the correctness of each segment ensures the validity of the entire path.
Conversely, if the overall reasoning path is determined to be incorrect and a specific intermediate step can be definitively identified as incorrect, all subsequent steps from that point onward are also labeled as incorrect, as errors propagate forward in the reasoning process. If no specific intermediate step can be identified as incorrect, the entire reasoning path is simply labeled as incorrect without making assumptions about the correctness or incorrectness of individual intermediate steps.

To diversify the PRM dataset, manual perturbations are applied to enrich it with diverse examples in both categories. Detailed descriptions of the PRM training dataset preparation are provided in Appendix~\ref{appendix:PRM_data_collection}. 

\section{Methodology}
\subsection{Training PRM}
\label{sec:methods:prm_inference}

% Previous works~\cite{openai-verifier-small-model,prm_google,prm_openai,mcts-3} have demonstrated that small-scale LLMs with verifiers that can evaluate the intermediate process can outperform foundational large-scale LLMs in mathematical reasoning tasks. In this work, we fine-tune Qwen2.5-Math-1.5B~\cite{qwen25-math} for a binary classification task. Details on constructing prompts for the PRM are provided in Appendix~\ref{appendix:prompt}. After fine-tuning, we extract the logits corresponding to the correct label and apply the sigmoid function to compute the score:
Previous works~\cite{openai-verifier-small-model,prm_google,prm_openai,mcts-3} have shown that small-scale LLMs equipped with verifiers evaluating intermediate processes are capable of outperforming foundational large-scale LLMs in mathematical reasoning tasks. In this work, we fine-tune Qwen2.5-Math-1.5B~\cite{qwen25-math} for a binary classification task. Details on constructing prompts for the PRM are provided in Appendix~\ref{appendix:prompt}. After full-parameter supervised fine-tuning, we extract the logits corresponding to the correct label and apply the sigmoid function to compute the score:

\begin{equation}
S_{\text{PRM}} = \frac{1}{1 + e^{-l_{prm}}},
\end{equation}
where $l_{prm}$ denotes the logit value for the correct label, and $S_{\text{PRM}}$ represents the PRM score.

\begin{figure}[t!]
    \centering
    \includegraphics[width=1\linewidth]{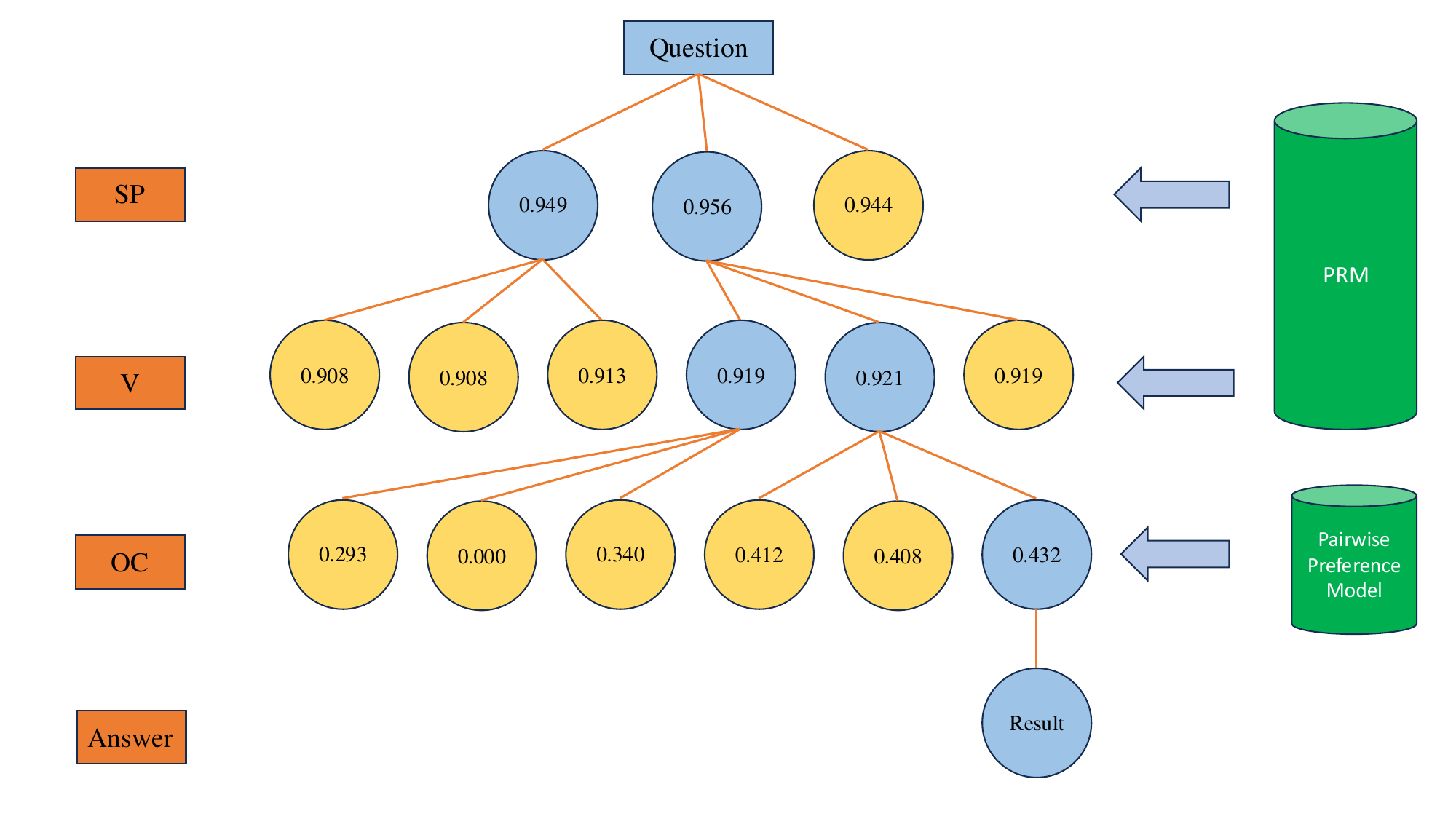}

    \caption{
A real demonstration of the BPP-Search process with a beam search width of 2. Yellow nodes represent pruned nodes that are not explored, while light blue nodes indicate nodes that have been visited.
}

    \label{fig:bpp}
\end{figure}

\subsection{BPP-Search}
\label{sec:method:bpp-search}

% We integrate the PRM with Greedy~\cite{greedy} and Beam Search~\cite{beam-search} algorithms, where the PRM provides scores to guide the selection of nodes to explore. However, as shown in Section~\ref{sec:experiment:ablation}, experimental results indicate that increasing the Beam Search width does not always improve performance and can sometimes degrade it. This issue likely arises because the PRM, trained for classification tasks, is required to assign continuous scores during inference. While the PRM performs well in distinguishing classes, it struggles to assign precise scores, leading to output discrepancies. Furthermore, manual analysis of the tree generation process reveals that candidates in the final layer of Beam Search sometimes include both correct and incorrect answers他们十分相似，细小区别，导致分数区别不大, with minimal score differences between them. This lack of significant score differentiation makes it challenging to identify the optimal answer effectively in the end.
We integrate the PRM with Greedy~\cite{greedy} and Beam Search~\cite{beam-search} algorithms, where the PRM provides scores to guide node selection. However, as shown in Section~\ref{sec:experiment:ablation}, increasing the Beam Search width does not consistently improve performance and, in some cases, leads to degradation. This limitation stems from the PRM, which is trained for classification tasks but required to assign continuous scores during inference, resulting in output discrepancies. Manual analysis of the tree generation process reveals that the final layer of Beam Search frequently contains both correct and incorrect candidates that are highly similar, with only subtle distinctions. These minor differences yield comparable scores, making it challenging to determine the optimal solution effectively.

To address this challenge, we propose \textbf{BPP-Search}, an algorithm that integrates \textbf{B}eam search, \textbf{P}RM, and a Pairwise \textbf{P}reference model. The core idea is to enhance decision-making in the final layer by leveraging a newly trained \textbf{Preference Model}, fine-tuned from Qwen2.5-Math-1.5B, to generate pairwise preference scores for ranking candidates.

\paragraph{Preference Model Data Preparation.} To train the preference model for pairwise preferences, we first use the ToT framework to generate a large set of unlabeled reasoning paths and classify them as correct or incorrect based on objective values. For each problem, one correct reasoning path (\(A\)) and one incorrect reasoning path (\(B\)) are extracted, and all pairwise combinations of correct and incorrect paths are generated. Prompts are constructed by arranging the problem with \(A\) and \(B\) in two different orders. If \(A\) appears first, the prompt is labeled as \(1\); otherwise, it is labeled as \(0\). The resulting labeled data is then used to fine-tune the Preference Model as a binary classifier. Details on constructing prompts for the Preference Model are provided in Appendix~\ref{appendix:prompt}.

\label{sec:methods:preference_data}

\paragraph{Pairwise Scoring and Candidate Ranking.} During inference, the preference model evaluates pairwise preference scores for any two candidates (\(A, B\)). The preference score \(S_{PM}(A  \succ B)\) is calculated as:
\begin{equation}
S_{PM}(A \succ B) = \frac{1}{1 + e^{-l_{pm}}},
\end{equation}
where \(l_{pm}\) is the logit value for class \(1\) and $A \succ B$ denotes the preference for $A$ over $B$.

To compute a comprehensive score for each candidate (\(A\)) in the candidate set, pairwise preference scores are aggregated across all other candidates:
% \begin{equation}
% S_{PM}(A) = \frac{1}{n} \sum_{j \neq i} S(A \succ X_j),
% \end{equation}
\begin{equation}
S_{PM}(A) = \frac{1}{n - 1} \sum_{j \neq i}^{j=1,\dots,n} S(A \succ X_j),
\end{equation}
% where \(n\) is the total number of candidates, \(X_i\) corresponds to \(A\) and \(X_j\) represents other candidates.
where \(n\) is the total number of candidates, \(i\) is the index corresponding to \(A\), and \(X_i\) represents \(A\), while \(X_j\) represents other candidates.

\begin{table*}[t]
\footnotesize
% \caption{
% Performance evaluation (number of correctly solved examples) of various LLMs on the StructuredOR test dataset (38 examples) under different methods: CoT \cite{cot} (including CoT-BMLD, where modeling is performed first and then data is imported, and CoT-SPVOC, which follows the sequence of set, parameter, variable, objective, and constraint for CoT), ToT \cite{tot}, SC \cite{sc}, ToT-randomly-chosen (where the final result is obtained randomly from the leaf nodes), ToT-rethink (providing all leaf nodes to the LLM and obtaining a revised result), and ToT-fully-traverse (where every leaf node is thoroughly checked).
% }
\centering
\resizebox{0.95\linewidth}{!}{
% \begin{tabular}{l|c|c|c}
\begin{tabular}{l|c|c|c|c|c|c}
\hline
\textbf{Model/Methods} & \textbf{CoT-BMLD} & \textbf{CoT-SPVOC} & \textbf{SC} & \textbf{ToT-Randomly-Chosen} & \textbf{ToT-Rethink} & \textbf{ToT-Fully-Traverse} \\
\hline
GPT-4o & \textbf{19} & \textbf{19} & \textbf{21} & \textbf{21} & \textbf{23} & \textbf{30}  \\
GPT-4o-mini & 13 & 14 & 19 & 12 & 19 & 21 \\
\hline
Llama-3-70B & 10 & 17 & 19 & 17 & 17 & 22 \\
Llama-3.1-70B & 18 & 10 & 21 & 5 & 14 & 17 \\
Llama-3.2-11B & 2 & 0 & 1 & 0 & 0 & 8 \\
\hline
Qwen-2-72B-Instruct & 2 & 2 & 4 & 3 & 1 & 5 \\
Qwen-2.5-MATH-72B-Instruct & 2 & 2 & 0 & 0 & 0 & 4\\
Qwen-2.5-72B-Instruct & 6 & 5 & 9 & 2 & 2 & 2\\
\hline
Mixtral-8$\times$7B-v0.1 & 0 & 0 & 0 & 0 & 1 & 1\\ 
\hline

\end{tabular}}

\caption{
Performance evaluation (number of correctly solved examples) of various LLMs on the StructuredOR test dataset (38 examples) under different methods: CoT \cite{cot} (including CoT-BMLD, where modeling is performed first and then data is imported, and CoT-SPVOC, which follows the sequence of set, parameter, variable, objective, and constraint for CoT), ToT \cite{tot}, SC \cite{sc}, ToT-randomly-chosen (where the final result is obtained randomly from the leaf nodes), ToT-rethink (providing all leaf nodes to the LLM and obtaining a revised result), and ToT-fully-traverse (where every leaf node is thoroughly checked).
}
\label{tab:baseline}

\end{table*}
\begin{table}[]
\footnotesize

\centering
\resizebox{\linewidth}{!}{
% \begin{tabular}{l|c|c|c}
\begin{tabular}{l|c}
\hline
\textbf{Dataset}          & \textbf{Number of Problems Resolved} \\ \hline
StructuredOR &30/38 \\ \hline
NL4OPT & 143/289 \\ \hline
ComplexLP & 72/211 \\ \hline
\end{tabular}}

\caption{
Number of problems resolved using ToT-Fully-Traverse. The numerator represents the number of examples with at least one correct answer, and the denominator indicates the total number of examples in the test dataset. Results are shown for GPT-4o on the StructuredOR, NL4OPT~\cite{nl4opt}, and ComplexLP~\cite{complexLP} datasets.
}
\label{tab:tot_fully_traverse}

\end{table}

\paragraph{Selection of the Optimal Candidate.} Using the computed scores \(S_{PM}(A)\), candidates are ranked, and the one with the highest score is selected as the optimal answer. This robust ranking mechanism addresses the limitations of the PRM, which struggles to differentiate between similar candidates and accurately identify the correct answer.

% Figure~\ref{fig:bpp} illustrates the BPP-Search process, demonstrating how nodes are pruned based on PRM scores during beam search (Width = 2) and 最后一层用pairwise preference algorithm找出最终结果. In the final layer, the remaining candidate answers are ranked using the Pairwise Preference Algorithm, and the optimal solution is selected, ensuring both robustness and accuracy in the final decision-making process.
Fig.~\ref{fig:bpp} illustrates the BPP-Search process, where nodes are pruned based on PRM scores during beam search (Width = 2), and the Pairwise Preference Algorithm ranks the final candidates to select the optimal solution, ensuring robustness and accuracy. Appendix~\ref{appendix:bpp_complexity} analyses computational cost of BPP-Search.

% \subsection{BPP Search Complexity Analysis}
% \label{sec:bpp_complexity}

% The time complexity for BPP-Search is 主要是$\mathcal{O}(n \cdot b \cdot (h-1) + n)$次policy model调用, where $h$ is the height of the tree, $b$ is the width of the beam search, and $n$ is the number of child nodes per parent node.

% The overall time complexity for BPP-Search consists of three components:
% \begin{enumerate}
%     \item $\mathcal{O}(n \cdot b \cdot (h-1) + n)$次policy model调用: The cost of accessing the policy model, where $h$ is the height of the tree, $b$ is the width of the beam search, and $n$ is the number of child nodes per parent node.
%     \item $\mathcal{O}(b^2 \cdot n^2)$次preference model调用: The cost of accessing the pairwise preference model to rank and select among candidates in the final layer.
%     \item $\mathcal{O}(n \cdot b \cdot (h-1))$次PRM调用: The cost of accessing the Process Reward Model (PRM) to evaluate intermediate steps during the search process.
% \end{enumerate}

\subsection{Random Greedy Algorithm}
\label{sec:method:random_greedy}

% To address the limitations of PRM's scoring precision, we propose the \textbf{Random Greedy} algorithm. Since PRM provides a rough preference ranking rather than precise scores, introducing randomness during the search process helps account for scoring uncertainties and mitigates the risk of suboptimal decisions.
To address the limitations of PRM's scoring precision, we employ the \textbf{Random Greedy} algorithm. Since PRM provides a rough preference ranking rather than precise scores, randomness is introduced to mitigate the impact of PRM's scoring variability.

The Random Greedy algorithm prioritizes candidates with scores close to the maximum while incorporating randomness to mitigate PRM's imprecision. Candidates are filtered based on the condition:
\begin{equation}
P(a_{\text{max}}) - P(a_i) \leq \text{threshold},
\end{equation}
where \(P(a_{\text{max}})\) is the highest score, \(P(a_i)\) is the score of candidate \(a_i\), and \(\text{threshold}\) is a predefined margin. From the filtered candidates, one is randomly selected to continue the search process.
\section{EXPERIMENT}

\subsection{Baseline}
\label{sec:experiemnt:baseline}

\begin{table*}[]
\footnotesize

\centering
\resizebox{0.95\linewidth}{!}{
% \begin{tabular}{l|c|c|c}
% \begin{tabular}{l|cr|cr|cr}
% \hline
% {\textbf{Method}} & \multicolumn{2}{c|}{\textbf{StructuredOR}} & \multicolumn{2}{c|}{\textbf{Mamo-ComplexLP}} & \multicolumn{2}{c}{\textbf{NL4OPT}} \\ 
%                                  & \textbf{Correct Rate} & \textbf{Steps}     & \textbf{Correct Rate} & \textbf{Steps}       & \textbf{Correct Rate} & \textbf{Steps}       \\ \hline
\begin{tabular}{l|c|r|c|r|c|r}
\hline
{\textbf{Method}} & \multicolumn{2}{|c|}{\textbf{StructuredOR}} & \multicolumn{2}{|c|}{\textbf{Mamo-ComplexLP}} & \multicolumn{2}{|c|}{\textbf{NL4OPT}} \\ \cline{2-7}
                                 & \textbf{Correct Rate} & \textbf{Steps}     & \textbf{Correct Rate} & \textbf{Steps}       & \textbf{Correct Rate} & \textbf{Steps}       \\ \hline

CoT~\cite{cot} & 0.633 & 1 & 0.486 & 1 & 0.566 & 1\\
SC~\cite{sc} & 0.700 & 4 & 0.625 & 4 & 0.713 & 4 \\
ToT-Randomly-Chosen~\cite{tot} & 0.700 & 39 & 0.444 & 39 & 0.629 &39\\
ToT-Rethink~\cite{tot} & 0.766 & 40 & 0.583 & 40 & 0.622 & 40\\ \hline

Greedy Search Variant (Our Method) & 0.833 & 9 & 0.555 & 9 & 0.713 & 9 \\ \hline
Beam Search Variant (Our Method) & 0.800 & 15 & 0.666 & 21 & 0.783 & 15 \\ \hline
\textbf{BPP-Search Variant (Our Method)} & \textbf{0.933} & 15 & \textbf{0.722} & 21 &  \textbf{0.804} & 15 \\ \hline
 
% Greedy & 0.733 & 9 & 0.555 &\\
% Randomly Greedy & 0.833 & 9 \\
% Epsilon Greedy & 0.733 & 9 \\ \hline
% Beam Search (Width=2) & 0.8 & 15 \\
% Beam Search (Width=3) & 0.766 & 21 \\ \hline
% Beam Search (Width=2) + PA & 0.933 & 15 \\
% Beam Search (Width=3) + PA & 0.866 & 21 \\
% \hline

\end{tabular}}

\caption{
Accuracy and reasoning steps for \textbf{BPP-Search} and baselines with a fixed policy model (GPT-4o) on the StructuredOR, Mamo-ComplexLP~\cite{complexLP}, and NL4OPT~\cite{nl4opt} test datasets. The results are based on 30 problems from StructuredOR, 72 from Mamo-ComplexLP, and 143 from NL4OPT that are confirmed solvable by the policy model in prior experiments.
}
\label{tab:our_methods}

\end{table*}
\begin{table*}[t]
\footnotesize

\centering
\resizebox{0.95\linewidth}{!}{
\begin{tabular}{l|c|c|c|c}
\hline
Method & StructuredOR Correct Rate & Mamo-ComplexLP Correct Rate & NL4OPT Correct Rate & Reasoning Step \\ \hline

Greedy Search + PRM & 0.733 & 0.555 & 0.699 & 9\\ 
Random Greedy Search + PRM & 0.833 & 0.513 & 0.692 & 9 \\
Epsilon Greedy Search + PRM & 0.733 & 0.500 & 0.713 & 9 \\ \hline

Beam Search (Width=2) + PRM & 0.800 & 0.652 & 0.783 & 15 \\ 
Beam Search (Width=3) + PRM & 0.766 & 0.666 & 0.755 & 21 \\ \hline

BPP-Search (Width=2) & \textbf{0.933} & 0.652 & \textbf{0.804} & 15 \\
BPP-Search (Width=3) & 0.866 & \textbf{0.722} & 0.797 & 21 \\ \hline

% Greedy & 0.733 & 9 & 0.555 &\\
% Randomly Greedy & 0.833 & 9 \\
% Epsilon Greedy & 0.733 & 9 \\ \hline
% Beam Search (Width=2) & 0.8 & 15 \\
% Beam Search (Width=3) & 0.766 & 21 \\ \hline
% Beam Search (Width=2) + PA & 0.933 & 15 \\
% Beam Search (Width=3) + PA & 0.866 & 21 \\
% \hline

\end{tabular}}

\caption{
Accuracy and reasoning steps for \textbf{ablation study} of our methods with a fixed policy model (GPT-4o) on the StructuredOR, Mamo-ComplexLP~\cite{complexLP}, and NL4OPT~\cite{nl4opt} test datasets. The results are based on the same 30 problems from StructuredOR, 72 from Mamo-ComplexLP, and 143 from NL4OPT, confirmed solvable by the policy model in prior experiments.
}
\label{tab:ablation}

\end{table*}

Given the varying performance levels of policy models across different scales, our objective is to maximize the accuracy of correct results by fully exploring every leaf node in the ToT structure, without requiring fine-tuning of the policy model.
Because this approach ensures greater stability and a larger pool of experimental data for subsequent analyses. 
To achieve this, we design a set of baseline experiments on the \textbf{StructuredOR}, providing more reliable and consistent evaluations.

We first evaluate the CoT~\cite{cot} approach, including two variations: CoT-BMLD, where the modeling process is performed first and data is imported later, and CoT-SPVOC, which adheres to the sequence of set, parameter, variable, objective, and constraint in the modeling process. 
Next, since the ToT~\cite{tot} framework lacks a mechanism to select a final answer from all leaf nodes, we introduce the following configurations to address this limitation: ToT-randomly-chosen, where the final result is randomly selected from the leaf nodes; ToT-rethink, where all leaf nodes are provided to the LLM for reevaluation to produce a revised result; and ToT-fully-traverse, where every leaf node is thoroughly evaluated to ensure that at least one correct result can be generated. The detailed tree structure is shown in the Fig.~\ref{fig:tot}.
 Additionally, we include SC~\cite{sc} as a baseline, which aims to obtain consistent results by sampling multiple reasoning paths.

The evaluated policy models include GPT-4o~\cite{gpt4}, GPT-4o-mini~\cite{gpt4}, Llama-3-70B~\cite{llama31}, Llama-3.1-70B~\cite{llama31}, Llama-3.2-11B~\cite{llama31}, Qwen-2-72B~\cite{qwen2}, Qwen-2.5-72B~\cite{qwen2.5}, Qwen-2.5-Math-72B~\cite{qwen25-math}, and Mixtral-7$\times$8B~\cite{mixtral}. Table~\ref{tab:baseline} shows the performance of baseline methods across these models in StructuredOR dataset. The results highlight significant variations in performance, demonstrating how model size and architecture impact their effectiveness in solving problems within the StructuredOR dataset.

To ensure stable performance in subsequent search algorithm experiments across different datasets, we select GPT-4o as the policy model. We first evaluate its ability to solve questions by finding at least one correct answer within the ToT fully-traverse framework on the test datasets from \textbf{StructuredOR}, \textbf{MAMO-ComplexLP}~\cite{complexLP}, and \textbf{NL4OPT}~\cite{nl4opt}. 
Table~\ref{tab:tot_fully_traverse} presents the results of this evaluation. For subsequent tree search algorithm experiments, only successful cases—where the policy model identifies at least one valid result—are considered. This ensures that the policy model can generate solutions for these questions within the ToT framework.

\subsection{Evaluation of BPP-Search}
\label{sec:experiment:bpp}

% We evaluate our methods, as mentioned in Section ~\ref{sec:method}, on the solvable problems identified in the datasets mentioned in Section~\ref{sec:experiemnt:baseline}. Table~\ref{tab:our_methods} compares our methods with baseline approaches in terms of correct rate and reasoning steps. These results demonstrate that, under the condition where none of the methods fine-tune the policy model, our methods achieve superior performance while requiring fewer reasoning steps, significantly outperforming SOTA baselines.
We evaluate our methods on the solvable problems identified in the datasets, as described in Section~\ref{sec:experiemnt:baseline}. 
We perform supervised fine-tuning of the PRM and Preference Model on their corresponding training datasets, as described in Sec.\ref{sec:method:prm_dataset} and Sec.\ref{sec:methods:preference_data}, respectively. Detailed evaluation results are provided in Appendix~\ref{appendix:PRM_RM_TRAIN}.
Table~\ref{tab:our_methods} presents a comparison between our methods and baseline approaches, focusing on correct rate and reasoning steps. The results show that, under the condition where none of the methods fine-tune the policy model, our methods achieve superior performance with fewer reasoning steps, significantly outperforming baselines.
% , as outlined in Section~\ref{sec:method:bpp-search}.

These experiments validate the feasibility of utilizing PRM to assist inference within the tree-of-thought structure in the domain of Operations Research. BPP-Search effectively addresses the limitations of traditional ToT methods, which struggle to reliably select a final result. As shown in Table~\ref{tab:our_methods}, Greedy Search, Beam Search, and BPP-Search generate better results in significantly fewer steps, exponentially reducing computational costs. 

% Despite the fine-tuned PRM's inability to assign precise scores, BPP-Search mitigates this issue through a pairwise preference scoring mechanism, which enhances result reliability and reduces the risks associated with imprecise scoring from the PRM.

\subsection{Ablation Study}
\label{sec:experiment:ablation}

% Table~\ref{tab:ablation} presents the performance of our methods under different configurations. It is evident that the PRM struggles to assign precise scores for regression tasks. For instance, as the beam search width increases, the accuracy tends to decrease, and in some cases, the performance of beam search becomes comparable to that of greedy search. Manual analysis of the beam search results in the final layer reveals that the candidate queue often contains both highly similar correct and incorrect answers, with minimal differences in their scores. This indicates that the PRM cannot reliably distinguish between candidates with close scores.
Table~\ref{tab:ablation} presents the performance of our methods under different configurations. It is evident that the PRM struggles to assign precise scores for regression tasks. For instance, as the beam search width increases, the accuracy tends to decrease, and in some cases, the performance of beam search becomes comparable to that of greedy search. Manual analysis of the beam search results in the final layer reveals that the candidate queue sometimes contains both correct and incorrect answers that are highly similar in structure, with only subtle differences. This similarity leads to comparable scores, making it challenging for the PRM to reliably distinguish between them.

To address this limitation, we introduce BPP-Search, which incorporates a pairwise preference algorithm. Instead of scoring candidates individually, BPP-Search evaluates all pairwise combinations of candidates within the final queue, comparing each pair and averaging the pairwise preference scores for each candidate. This approach ensures a more robust evaluation by reducing the bias inherent in relying solely on individual scores. The experimental results in Table~\ref{tab:ablation} demonstrate that this method effectively mitigates the risks associated with the imprecise scoring of the PRM, resulting in improved accuracy and robustness.
Additionally, based on the performance of PRM, our random greedy search algorithms can at least guarantee performance comparable to standard greedy search, and in some cases, achieve even better results.

% \input{tables/structuredOR}
% \input{tables/nl4opt}
% \input{tables/complex_lp}
% \section{Conclusion}
% In this work, we address the limitations of existing open-source operations research datasets by releasing a new dataset that includes both natural language questions and their corresponding detailed modeling processes, filling the gap in current resources that lack such annotations.
% %
% We further propose BPP-Search, a novel algorithm that integrates Beam Search, PRM, and a Pairwise Preference mechanism. BPP-Search is designed to enhance the ToT framework by accelerating the inference process and improving the accuracy of final results. BPP-Search effectively accelerate reasoning process and improve accuracy, addressing the scoring imprecision of PRM and ensuring robust decision-making.
% %
% Extensive experiments conducted on StructuredOR, NL4OPT, and MAMO-ComplexLP datasets demonstrate the superiority of BPP-Search. Compared to state-of-the-art baselines, including CoT, SC, and PRM combined with Greedy or Beam Search, our approach achieves higher accuracy while requiring fewer reasoning steps.
\section{Conclusion}
In this work, we introduce a new operations research dataset that integrates natural language questions with their corresponding detailed modeling processes, addressing the limitations of existing open-source datasets that lack comprehensive annotations of the modeling process.
We further propose BPP-Search, an advanced algorithm that combines Beam Search, PRM, and a Pairwise Preference mechanism to enhance the ToT framework. BPP-Search effectively accelerates the reasoning process, improves accuracy, and alleviates the scoring imprecision of PRM, thereby ensuring robust and reliable decision-making.
Comprehensive experiments conducted on StructuredOR, NL4OPT, and MAMO-ComplexLP datasets highlight the superiority of BPP-Search. Compared to state-of-the-art approaches, such as CoT, SC, and PRM integrated with Greedy or Beam Search, BPP-Search consistently achieves higher accuracy while requiring fewer reasoning steps, demonstrating its efficacy in addressing complex reasoning tasks in operations research.

% \section{Limitations}
% \subsection{Limitations of ToT Structure: Width and Depth}
% In the ToT framework, both increasing tree width and deepening the tree can enhance performance. A wider tree, achieved by increasing the number of child nodes at each layer, provides more exploration paths, improving the likelihood of finding optimal solutions. On the other hand, greater depth, achieved by dividing tasks into finer-grained nodes (e.g., separating "set" and "parameter" into distinct layers), not only facilitates more structured and detailed reasoning but also provides additional exploration paths, thereby further increasing the likelihood of identifying optimal solutions.

% However, both approaches come with significant computational costs. For instance, a tree with a height of 4 and a branching factor of 3 requires 39 LLM queries, while increasing the branching factor to 4 raises this to 84 queries, thereby exacerbating computational costs, particularly for long prompts (e.g., 4000 tokens). This creates a trade-off between computational cost and performance, necessitating the selection of appropriate tree parameters within the constraints of available computational resources and budget.
\section{Limitations}
\subsection{Trade-offs in ToT Structure: Performance and Computational Cost}
In the ToT framework, both increasing tree width and deepening the tree can enhance performance. A wider tree, achieved by increasing the number of child nodes at each layer, provides more exploration paths, thereby improving the likelihood of finding optimal solutions. Similarly, greater depth, achieved by dividing tasks into finer-grained nodes (e.g., separating "set" and "parameter" into distinct layers), not only facilitates more structured and detailed reasoning but also offers additional exploration paths, further increasing the likelihood of identifying optimal solutions.

However, both approaches entail significant computational costs. For example, a tree with a height of 4 and a branching factor of 3 requires 39 LLM queries, whereas increasing the branching factor to 4 raises this to 84 LLM queries, thereby exacerbating computational demands, particularly for long prompts (e.g., 4000 tokens). This creates a trade-off between computational cost and performance. In our study, due to limitations in computational resources, we were unable to construct a tree that is both sufficiently deep and wide to fully explore the solution space.

% \subsection{Dillemma between LLM Capability and OR problem complexity}
% In operations research, the problem can be generalized to 几种类别，which can be considered as the "seed" data. The reason why we don't expand StructuredOR dataset scale is that the specifics of OR problems is limited and if we just replace the value of some parameters these problem is highly similar, which leads many issue, such as test data leap, overfit, etc. 

% But if just expand the scale of OR problem, there  will be tons of number in the question, which also leads a question that current LLM cannot memorize numberous number and output them orderly. The key limitation lies in the current capabilities of LLMs, which struggle to handle large parameter spaces in an orderly manner. For example, a parameter representing transport costs, Cost[X][A][t], with dimensions (X), (A), and (t) containing 2 items each results in (2^3 = 8) values; with 3 items per dimension, this grows to (3^3 = 27). Such complexity increases exponentially with additional dimensions. At this stage, LLMs are not capable of processing large-scale parameters in an orderly way. Therefore, we focus on reasoning abilities rather than memorization. To ensure meaningful and feasible data, we avoid casual augmentation and carefully select examples aligned with the model's capabilities.

\subsection{Dilemma between LLM Capability and OR Problem Complexity}

In operations research, most problems can be categorized into a limited number of canonical types, which we treat as the "seed" problems. We deliberately avoid indiscriminate expansion of the StructuredOR dataset because simply modifying parameter values results in highly similar instances. This introduces risks such as test set leakage and model overfitting, reducing the robustness of experimental evaluations.

On the other hand, if we were to diversify the dataset by significantly increasing the size and dimensionality of the problem instances, the resulting questions would contain a large number of numerical values. Current LLMs face fundamental limitations in this regard, as they struggle to memorize and accurately reproduce numerous numerical inputs in a structured and orderly fashion.

For instance, a single parameter such as $\texttt{Cost}[X][A][t]$ with three dimensions—each containing 3 elements—already results in $3^3 = 27$ values. As the number of elements in each dimension increases, the parameter space grows exponentially. While this level of complexity is trivial for mathematical solvers, it poses a significant challenge for LLMs, which are not yet capable of accurately reproducing large-scale numerical values in a complete and orderly manner.

\newpage
% \clearpage
\bibliography{ref}

\newpage
\appendix

\section{Appendix}

\subsection{Modeling Data Format Specification}
\label{appendix:convention}
Tables ~\ref{tab:abstract_modeling},~\ref{tab:set_component},~\ref{tab:parameter_component},~\ref{tab:variable_component},~\ref{tab:objective_component},~\ref{tab:constraint_component},~\ref{tab:formula_details},~\ref{tab:unsupported_formulas},~\ref{tab:supported_special_formulas} define the standardized data format for representing mathematical models. This convention provides a structured and consistent way to label and organize modeling data, ensuring clarity and usability across different tasks and datasets.
\begin{table*}[t!]
\centering
\resizebox{\linewidth}{!}{
\begin{tabular}{l|l|l|l}
\hline
\textbf{Name}       & \textbf{Type} & \textbf{Required} & \textbf{Description}                                  \\ \hline
set                 & list          & No                & Collection definitions needed for abstract modeling   \\ \hline
parameter           & list          & No                & Constants needed for abstract modeling               \\ \hline
variable            & list          & Yes               & Variables required for modeling                      \\ \hline
objective           & list          & Yes               & Objective function definition required for modeling  \\ \hline
constraint          & list          & No                & Constraints needed for modeling                      \\ \hline
\end{tabular}}
\caption{Summary of components for abstract modeling}
\label{tab:abstract_modeling}
\end{table*}

\begin{table*}[t!]
\centering

\begin{tabular}{p{2.5cm}|p{2cm}|p{10cm}} % Use curly braces {} with p for specifying column widths
\hline
\textbf{Name}       & \textbf{Type} & \textbf{Description}                                                                                     \\ \hline
name                & str           & Collection name, must meet programming naming conventions, no spaces allowed                             \\ \hline
description         & str           & Description of the collection                                                                            \\ \hline
data                & list          & Use a list starting from 1 and ending at the size of the set to represent the number of elements in the set. \\ \hline
\end{tabular}
\caption{Details of the Set Component}
\label{tab:set_component}
\end{table*}

\begin{table*}[t!]
\centering
\resizebox{\linewidth}{!}{
\begin{tabular}{p{2.5cm}|p{2cm}|p{10cm}}\hline
\textbf{Name}       & \textbf{Type} & \textbf{Description}                                                                                                                           \\ \hline
name                & str           & Parameter name, must meet programming naming conventions, no spaces allowed                                                                    \\ \hline
description         & str           & Parameter description                                                                                                                          \\ \hline
domain              & str           & The index dimension of the parameter, e.g., “{a \textless in\textgreater Aircraft}”. If this parameter is a constant, the domain is an empty string. If this parameter is multi-dimensional, please list the corresponding index. \\ \hline
data                & list          & Use a list or a number depending on whether the domain is an empty string. If the domain is an empty string, the data is a number. Otherwise, it is a list that can be either one-dimensional or multi-dimensional, representing the values of each parameter across different sets. There is a one-to-one correspondence between the dimensions of data and the domain. \\ \hline
\end{tabular}}
\caption{Details of the Parameter Component}
\label{tab:parameter_component}
\end{table*}

\begin{table*}[t!]
\centering
\resizebox{\linewidth}{!}{

\begin{tabular}{p{2.5cm}|p{2cm}|p{10cm}}\hline
\textbf{Name}       & \textbf{Type} & \textbf{Description}                                                                                     \\ \hline
name                & str           & Variable name, must meet programming naming conventions, no spaces allowed                              \\ \hline
description         & str           & Variable description                                                                                     \\ \hline
domain              & str           & Index dimension of the variable, e.g., ``{a \textless in\textgreater Aircraft}"                          \\ \hline
type                & str           & Variable type: CONTINUOUS, INTEGER, BINARY. Default is CONTINUOUS. Case insensitive                      \\ \hline
\end{tabular}}
\caption{Details of the Variable Component}
\label{tab:variable_component}
\end{table*}

\begin{table*}[t!]
\centering
\resizebox{\linewidth}{!}{
\begin{tabular}{p{2.5cm}|p{2cm}|p{10cm}}\hline
\textbf{Name}       & \textbf{Type} & \textbf{Description}                                                                                     \\ \hline
name                & str           & Objective function name, must meet programming naming conventions, no spaces allowed                     \\ \hline
description         & str           & Objective function description                                                                           \\ \hline
sense               & str           & Optimization direction of the objective function: min, max, minimize, maximize                           \\ \hline
function            & str           & Formula of the objective function, details below                                                        \\ \hline
\end{tabular}}
\caption{Details of the Objective Component}
\label{tab:objective_component}
\end{table*}

\begin{table*}[t!]
\centering
\begin{tabular}{p{2.5cm}|p{2cm}|p{10cm}}\hline
\textbf{Name}       & \textbf{Type} & \textbf{Description}                                                                                     \\ \hline
name                & str           & Constraint name, must meet programming naming conventions, no spaces allowed                             \\ \hline
description         & str           & Constraint description                                                                                   \\ \hline
domain              & str           & Index dimension of the constraint. Without filter: ``{a \textless in\textgreater Aircraft}".             \\ \hline
function            & str           & Formula of the constraint, details below                                                                 \\ \hline
\end{tabular}
\caption{Details of the Constraint Component}
\label{tab:constraint_component}
\end{table*}

\begin{table*}[t!]
\centering
\resizebox{\linewidth}{!}{

\begin{tabular}{p{2.5cm}|p{4cm}|p{8cm}}\hline
\textbf{Type}                  & \textbf{Expression}                  & \textbf{Description}                                              \\ \hline
Formula with sum symbol        & $\sum_{i \in I} x_{i}$               & $\{i \in I\}$ is the summation dimension                          \\ \hline
\end{tabular}}
\caption{Details of Formula Expressions}
\label{tab:formula_details}
\end{table*}

\begin{table*}[t!]
\centering
\resizebox{\linewidth}{!}{

\begin{tabular}{p{2.5cm}|p{4cm}|p{8cm}}\hline
\textbf{Type}                       & \textbf{Expression}                                 & \textbf{Description}                                                             \\ \hline
Sum dimension with subscript parameter & $\sum_{i \in \text{Successors}_{k}}$               & Does not support $\text{Successors}_{k}$ forms                                  \\ \hline
Nested parentheses in index dimension & $\{i \in P \{k \in A \{l \in A\}\} \in \text{NOE}_{i}\}$ & Nested parentheses in index dimension are not supported                         \\ \hline
Subscript restriction                 & $x_{i,j}$ where $i, j$ cannot be a number           & Numeric subscripts like $x_{i,1}$, $x_{1,j}$ are not supported                  \\ \hline
\end{tabular}}
\caption{Unsupported Formulas and Their Limitations}
\label{tab:unsupported_formulas}
\end{table*}

\begin{table*}[t!]
\centering
\begin{tabular}{p{2.5cm}|p{4cm}|p{8cm}}
\hline
\textbf{Type}                        & \textbf{Expression}                             & \textbf{Description}                                                                                     \\ \hline
Continuous expression                & $a < b < c$                                    & Continuous expressions are supported.                                                                   \\ \hline
Expression separated by commas       & $x + y < 0, \, y + z < 1$                      & Must be separated by English commas; will be split into two constraints.                                \\ \hline
Two consecutive sums                 & $\sum_{i \in I} \sum_{j \in J} x_{ij}$         & Will be merged into: $\sum_{i \in I, j \in J} x_{ij}$.                                                  \\ \hline
Missing * multiplication symbol      & $\sum_{i \in I} (a_{i} x_{i} + b_{i} y_{i})$   & Multiplication symbol will be automatically filled.                                                     \\ \hline
\end{tabular}
\caption{Supported Special Formulas}
\label{tab:supported_special_formulas}
\end{table*}

% \onecolumn
% \input{imgs/convention}
% \twocolumn

\subsection{Operations Research Dataset Comparison}
\label{appendix:OR_dataset}

\begin{figure*}[t!]
    \centering
    \includegraphics[width=0.9\linewidth]{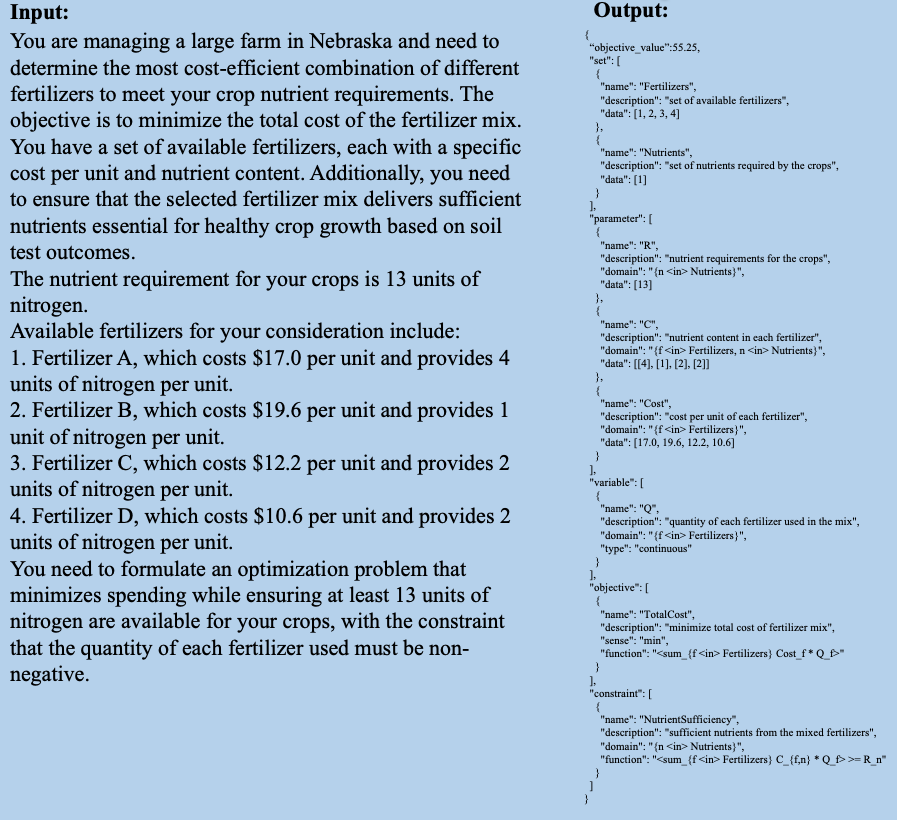}
    \caption{An example showcasing a concrete question and its structured modeling process as the label in the StructuredOR dataset.}
    \label{fig:example_structure}
\end{figure*}

\begin{figure*}[t!]
    \centering
    \includegraphics[height=0.5\linewidth]{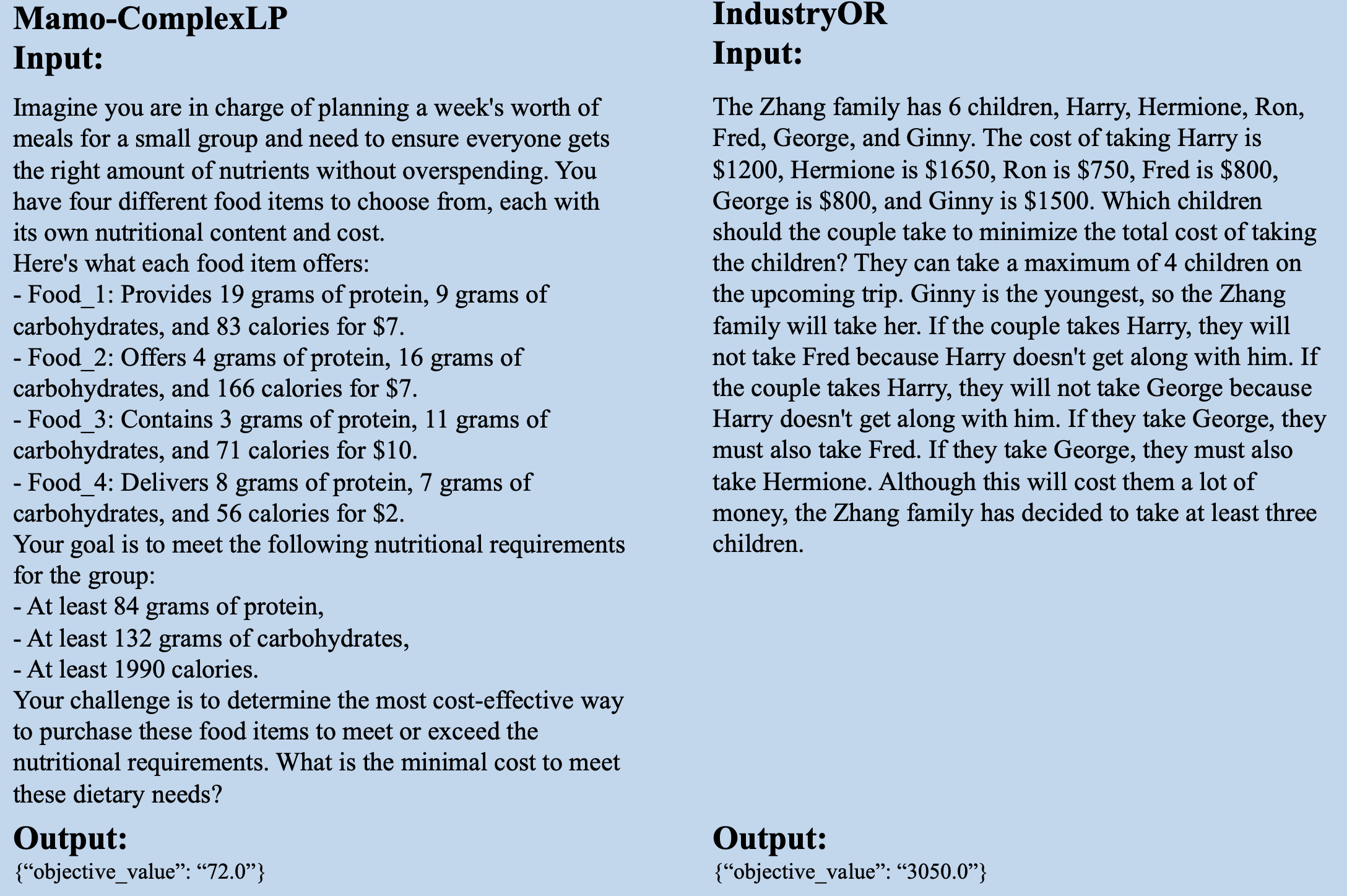}
    \caption{Comparison highlighting the limitations of the Mamo-ComplexLP and IndustryOR datasets.}
    \label{fig:other_dataset}
\end{figure*}

The StructuredOR dataset, as illustrated in Figure~\ref{fig:example_structure}, provides not only the objective value but also the complete modeling process, offering a structured and transparent view of optimization problems. In contrast, as shown in Figure~\ref{fig:other_dataset}, the Mamo-ComplexLP~\cite{complexLP} and IndustryOR~\cite{industryOR} datasets include only the objective value as the label, without detailing the modeling process. This limitation makes it difficult to verify the correctness of the data and prevents the application of reinforcement learning to intermediate steps. Similarly, the NL4OPT~\cite{nl4opt} dataset lacks both a structured modeling process and clear objective values, further complicating the interpretation and validation of results.

\subsection{An Example to Illustrate Reasoning in Mathematical Modeling}
\label{appendix:reasoning_example}
\begin{figure*}[t!]
    \centering
    \includegraphics[width=1\linewidth]{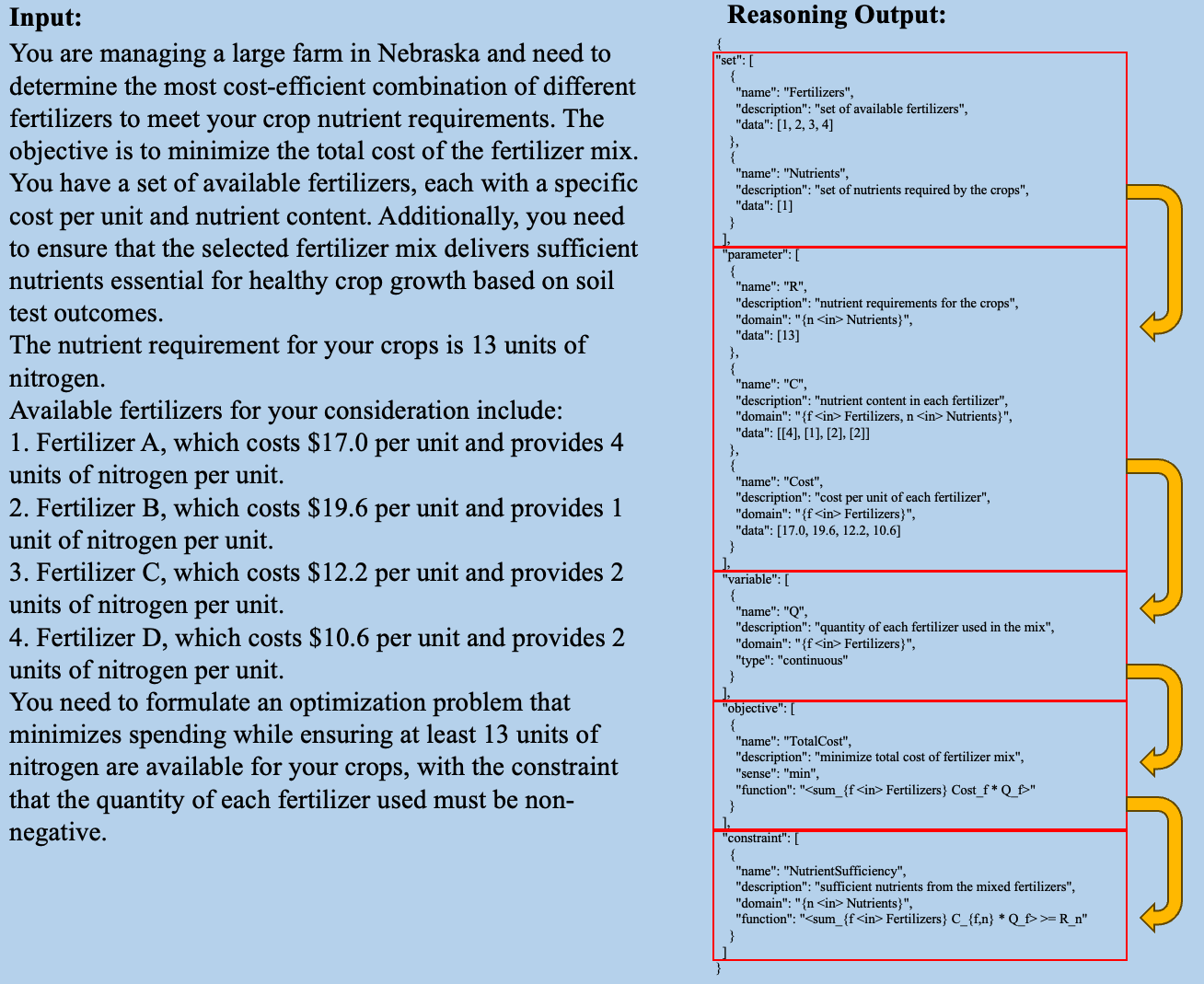}
    \caption{An example illustrating the reasoning process in mathematical modeling.}
    \label{fig:reasoning_example}
\end{figure*}

% Figure~\ref{fig:reasoning_example} provides an example of the reasoning steps in X-of-Thought (XoT) framework under the convention of modeling strcutre定义的格式~\ref{appendix:convention}. The process follows a structured sequence: starting with the question (\(Q\)), followed by sets (\(S\)), parameters (\(P\)), variables (\(V\)), objectives (\(O\)), and finally constraints (\(C\)). Each step builds upon the previous one, progressively transforming the natural language question into a fully defined mathematical model.
Figure~\ref{fig:reasoning_example} illustrates the reasoning steps in the X-of-Thought (XoT) framework under the modeling structure defined in Appendix~\ref{appendix:convention}. The process follows a structured sequence, starting with the question (\(Q\)), then progressing through sets (\(S\)), parameters (\(P\)), variables (\(V\)), objectives (\(O\)), and constraints (\(C\)). Each step builds upon the previous one, progressively transforming the natural language question into a fully defined mathematical model.

\subsection{PRM Training Data Collection}
\label{appendix:PRM_data_collection}
To augment positive data for Process Reward Model training, we adopt the following four strategies:

\begin{enumerate}
    \item \textbf{Utilizing ground truth:} Segment the ground truth data into accumulative chunks corresponding to different layers of the reasoning process. This ensures that the hierarchical structure of the data is preserved.

  \item \textbf{Leveraging LLM-generated data:} Identify correctly generated data from LLMs operating under ToT, CoT, and SC frameworks, and apply the same segmentation operations used for correct generated data. This approach expands the dataset with additional examples while ensuring consistency and alignment with the hierarchical structure.

  \item \textbf{Swapping indices in summation constraints:} Exchange indices within summation functions in constraints derived from ground truth data. This operation does not alter the final result, thereby introducing diversity while preserving correctness.

\item \textbf{Modifying inequalities:} Swap the left-hand and right-hand sides of inequalities derived from ground truth data, and adjust the inequality signs accordingly (e.g., `>=` becomes `<=`). This operation creates valid variations of the data while maintaining correctness.

\end{enumerate}

To augment incorrect data for Process Reward Model training, we apply the following strategies:

\begin{enumerate}
    \item \textbf{Mismatch instance data:} Replace the correct instance data with mismatched values. For example:
    \begin{itemize}
        \item Modify the value of a parameter so that it no longer corresponds to the data of the set.
        \item Delete or add random data to a `set`.
        \item Delete a column from a random dimension of a parameter.
        \item Reshuffle the data of a random parameter.
    \end{itemize}

    \item \textbf{Incorrect format:} Generate data using LLMs based on the training dataset, then select examples that cannot be used for modeling due to structural inconsistencies or formatting issues. 

    \item \textbf{Constraint modifications:} Introduce errors in constraints or objectives by:
    \begin{itemize}
        \item Changing a greater-than sign into a less-than sign.
        \item Swapping the indices within a constraint.
        \item Altering the summation domain of a constraint.
        \item Randomly deleting a constraint. 
        \item Modifying the function in either the constraint or the objective.
    \end{itemize}

    \item \textbf{Objective reversals:} Convert a minimization objective into its maximization counterpart, or vice versa.

\item \textbf{Generated incorrect models:} Utilize LLM-generated data that is structurally valid and adheres to modeling conventions but produces incorrect results, where the objective value from the modeling solution deviates from the expected outcome. This approach ensures the data aligns with modeling principles while intentionally introducing errors in reasoning or optimization outcomes.

\end{enumerate}

\subsection{PRM and Preference Model Prompt}
\label{appendix:prompt}
\begin{figure*}[t!]
    \centering
    \includegraphics[width=0.9\linewidth]{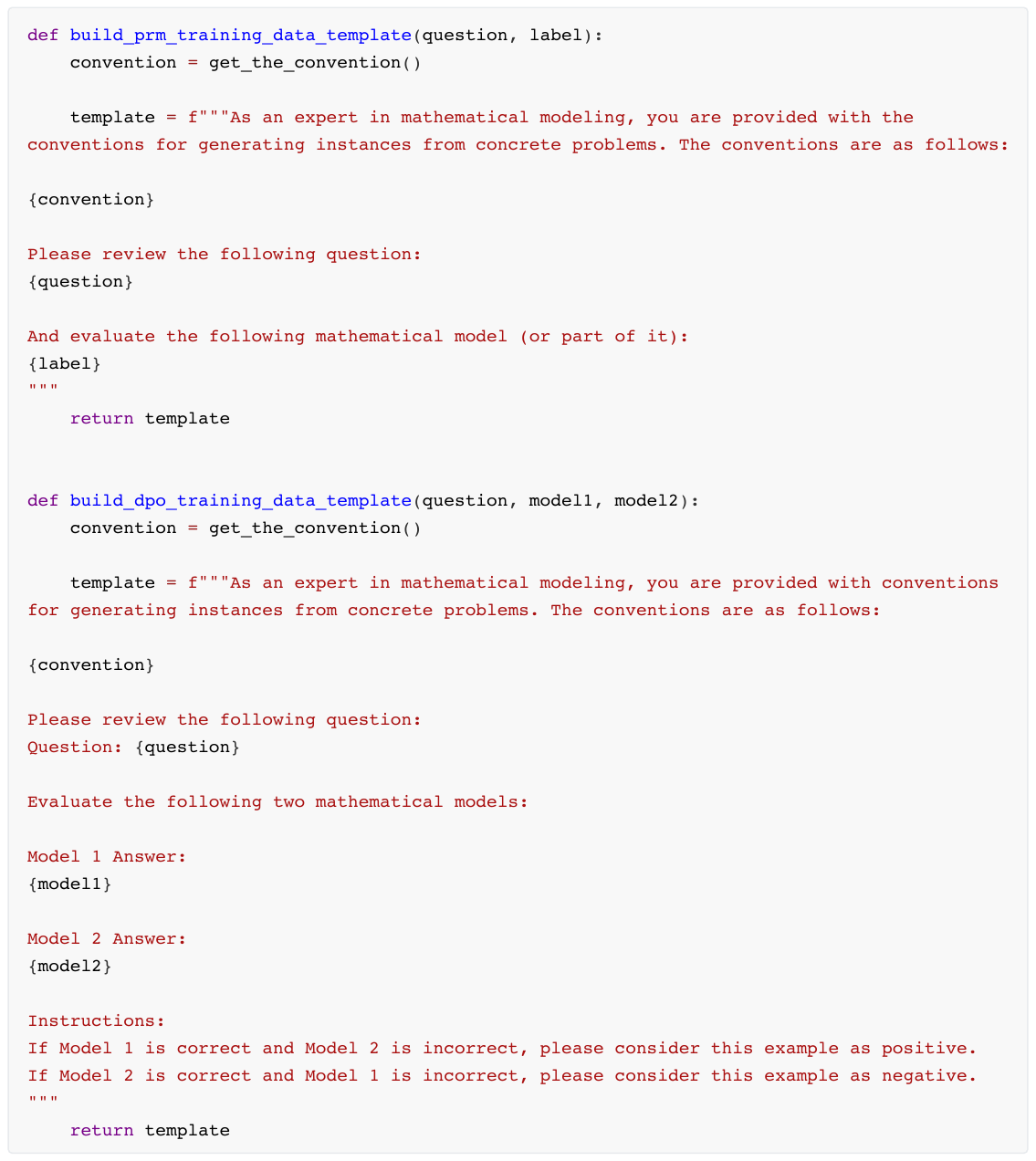}
    \caption{How PRM and the Preference Model construct prompts.}
    \label{fig:prm_prompt}
\end{figure*}

Figure~\ref{fig:prm_prompt} shows two functions illustrating how PRM and Preference Model construct prompts.

\subsection{BPP Search Complexity Analysis}
\label{appendix:bpp_complexity}

The computational complexity of BPP-Search is determined by the number of invocations of the policy model, the pairwise preference model, and the PRM. Specifically, the overall complexity can be decomposed into the following three components:

Here, $h$ denotes the height of the search tree, $b$ represents the beam width, and $n$ is the number of child nodes per parent node. 

\paragraph{Policy Model Invocations:} $\mathcal{O}(n \cdot b \cdot (h{-}1) + n)$. This term quantifies the cost incurred by the policy model during the node expansion process across the tree levels.
    
\paragraph{Pairwise Preference Model Invocations:} $\mathcal{O}(b^2 \cdot n^2)$. This cost arises from ranking and selecting candidates in the final layer, where each candidate is compared pairwise with the others.
    
\paragraph{PRM Invocations:} $\mathcal{O}(n \cdot b \cdot (h{-}2) + n)$. This component reflects the number of calls to the PRM for evaluating the intermediate steps throughout the search.

Since the reward model and the pairwise preference model, typically implemented with 1.5B or 3B parameters, are relatively small compared to the policy model (usually a 70B, 405B parameter model), the primary concern is the time spent calling the policy model.

\subsection{PRM, PM Training}
\label{appendix:PRM_RM_TRAIN}

Table~\ref{tab:prm_performance} and Table~\ref{tab:preference_model} report the performance of the PRM and the Preference Model on their corresponding test sets after supervised fine-tuning.

\begin{table}[t]
\footnotesize
\centering

\resizebox{\linewidth}{!}{
\begin{tabular}{l|c|c|c|c}
\hline
\textbf{Metric}    & \textbf{Accuracy} & \textbf{Precision} & \textbf{Recall} & \textbf{F1-Score} \\ \hline
\textbf{Value}     & 0.9823            & 0.9772             & 0.9868          & 0.9820            \\ \hline
\end{tabular}}

\caption{Performance metrics of the PRM on the PRM test dataset (Section~\ref{sec:method:prm_dataset}).}

\label{tab:prm_performance}

\end{table}

\begin{table}[t]
\footnotesize
\centering

\resizebox{\linewidth}{!}{
\begin{tabular}{l|c|c|c|c}
\hline
\textbf{Metric}    & \textbf{Accuracy} & \textbf{Precision} & \textbf{Recall} & \textbf{F1-Score} \\ \hline
\textbf{Value}     & 0.7560            & 0.7761             & 0.7196          & 0.7468            \\ \hline
\end{tabular}}

\caption{Performance metrics of the Preference Model on the Preference Model test dataset (Section~\ref{sec:methods:preference_data}) }
\label{tab:preference_model}

\end{table}

\end{document}